\documentclass[times,twocolumn,final,authoryear]{elsarticle}

\usepackage{ycviu}
\usepackage{framed,multirow}

\usepackage{amssymb}
\usepackage{latexsym}

\usepackage{url}
\usepackage{xcolor}
\usepackage{graphicx}
\usepackage{comment}
\usepackage{amsmath} 
\usepackage{color}
\usepackage{xspace}
\usepackage{subfigure}
\usepackage{booktabs}
\usepackage{bm}
\usepackage{enumitem}
\definecolor{newcolor}{rgb}{.8,.349,.1}

\def \ie {\emph{i.e.}}

\def \ours {STAGE\xspace}

\newcommand{\tit}[1]{\smallbreak\noindent\textbf{#1.}}

\begin{document}

\thispagestyle{empty}

\begin{frontmatter}

\title{Video action detection by learning graph-based spatio-temporal interactions}

\author[1]{Matteo \snm{Tomei}\corref{cor1}}
\cortext[cor1]{Corresponding author: }
\ead{matteo.tomei@unimore.it}
\author[1]{Lorenzo \snm{Baraldi}}
\author[1]{Simone \snm{Calderara}}
\author[2]{Simone \snm{Bronzin}}
\author[1]{Rita \snm{Cucchiara}}

\address[1]{University of Modena and Reggio Emilia, via Pietro Vivarelli 10, Modena 41125, Italy}
\address[2]{METALIQUID S.R.L., Via Giosue Carducci 26, Milano 20123, Italy}

\received{1 May 2013}
\finalform{10 May 2013}
\accepted{13 May 2013}
\availableonline{15 May 2013}
\communicated{S. Sarkar}

\begin{abstract}
Action Detection is a complex task that aims to detect and classify human actions in video clips. Typically, it has been addressed by processing fine-grained features extracted from a video classification backbone. Recently, thanks to the robustness of object and people detectors, a deeper focus has been added on relationship modelling. Following this line, we propose a graph-based framework to learn high-level interactions between people and objects, in both space and time.
In our formulation, spatio-temporal relationships are learned through self-attention on a multi-layer graph structure which can connect entities from consecutive clips, thus considering long-range spatial and temporal dependencies.
The proposed module is backbone independent by design and does not require end-to-end training. Extensive experiments are conducted on the AVA dataset, where our model demonstrates state-of-the-art results and consistent improvements over baselines built with different backbones. \textcolor{black}{Code is publicly available at \url{https://github.com/aimagelab/STAGE_action_detection}}.
\end{abstract}

\begin{keyword}
\MSC 41A05\sep 41A10\sep 65D05\sep 65D17
\KWD Keyword1\sep Keyword2\sep Keyword3

\end{keyword}

\end{frontmatter}



\section{Introduction}
\label{sec:introduction}
\begin{figure}[t]
    \centering
    \includegraphics[width=.95\linewidth]{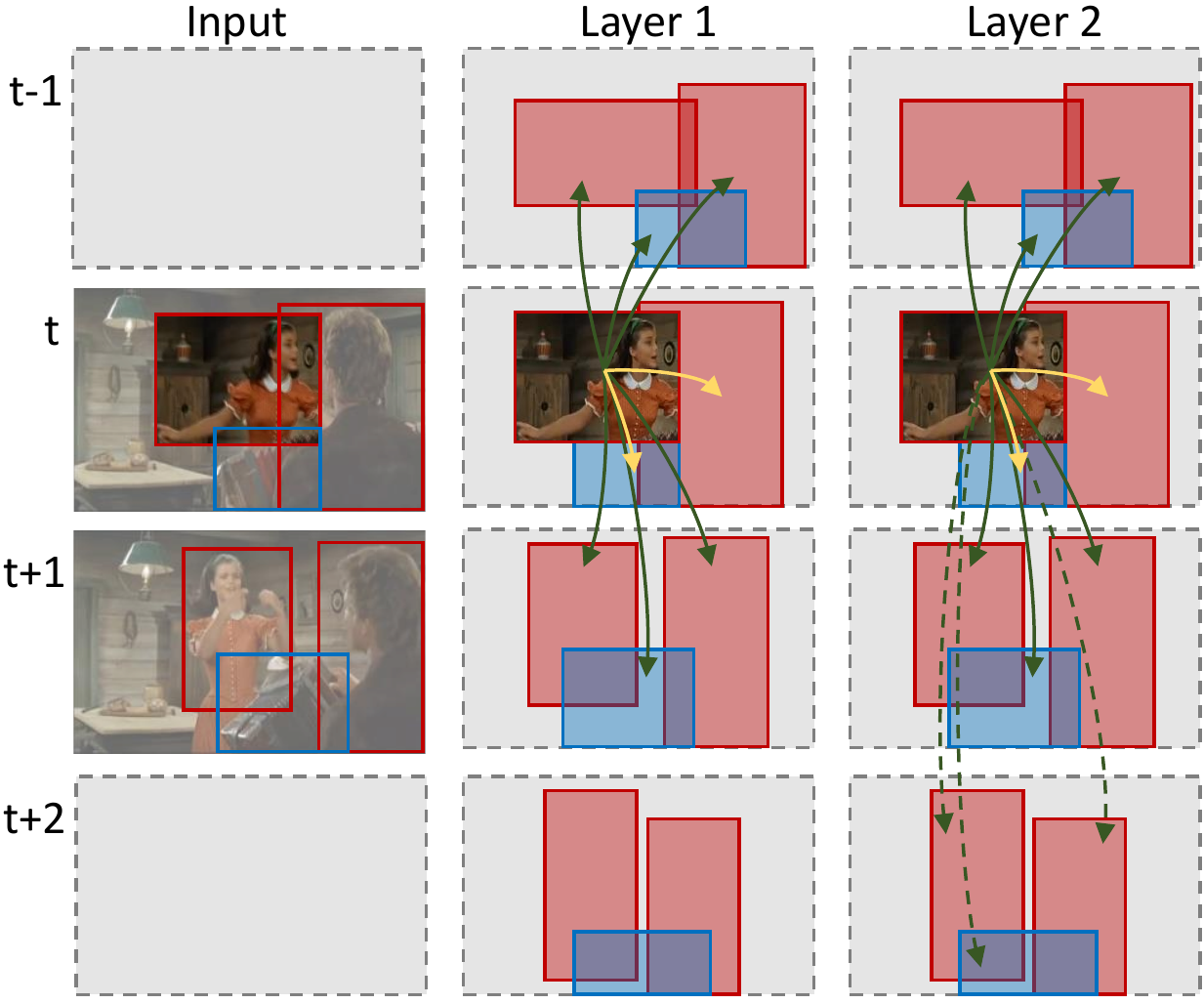}
    \caption{\textcolor{black}{We propose a graph-based module for video action detection which encodes relationships between actors and objects in a spatio-temporal neighborhood. Multiple layers of the module generate indirect edges between temporally distant entities, increasing the temporal receptive field.}}
    \label{fig:graph}
    \vspace{-.3cm}
\end{figure}

\textcolor{black}{Understanding people actions in video clips is an open problem in computer vision, which has been addressed for more than twenty years ~\citep{bobick2001recognition,herath2017going}. In the past, this task was tackled through handcrafted features designed for specific actions~\citep{laptev2005space,vezzani2009efficient}}. Recently, the video action detection task~\citep{sun2018actor,ulutan2020actor,yang2019step} was introduced along with deep architectures able to extract fine-grained and discriminative spatio-temporal features, to represent video chunks in a compact and manageable form. This has motivated recent efforts to design novel backbones for video feature extraction~\citep{feichtenhofer2019slowfast,tran2018closer,wu2019long}. On the other hand, higher-level reasoning is necessary for detecting and understanding human actions.

\textcolor{black}{
Interestingly, the performances of video action detection networks that take inspiration from object detection architectures are still far from being satisfactory. For example, it would be difficult to recognize whether a person is \textit{watching} someone just by looking at a bounding box around him, without considering the context. This can be partly explained by the lack of proper context understanding of the previous works, as they cannot model the relationships between actors and surrounding elements~\citep{ulutan2020actor}. Also, the presence of objects and other people in the scene, together with their behaviors, influences the understanding of the actor at hand. 
}

High-level reasoning is necessary not only at the spatial level, to model relations between close entities, but also in time: most of the existing backbones can handle small temporal variations, without modeling long-term temporal relationships.

\textcolor{black}{Following these premises, we devise a high-level module for video action detection which considers interactions between different people in the scene and interactions between actors and objects. Further, it can also take into account temporal dependencies by connecting consecutive clips during learning and inference. 
The same module can be stacked multiple times to form a multi-layer structure (Fig.~\ref{fig:graph}). In this manner, the overall temporal receptive field can be arbitrarily increased to model long-range dependencies. 
Since our method works at the feature level, it can expand its temporal receptive field without dramatically increasing its computational requirements. Our solution can exploit existing backbones for feature extraction and can achieve state-of-the-art results without an end-to-end finetuning of the underlying backbone.}

Previous works in action analysis have already tried to exploit graph-based representations~\citep{wang2018videos,zhang2019structured}, to model relationships with the context~\citep{girdhar2019video,ulutan2020actor} and to exploit long-term temporal relations~\citep{wu2019long}: our proposal merges all these insights in a single module, which is independent of the feature extraction layers and works on pre-computed representations. Moreover, our model is the first to employ a learning-based approach also on graph edges. \textcolor{black}{We test our model on the Atomic Visual Actions (AVA) dataset~\citep{gu2018ava}, which represents a challenging test-bed for recognizing human actions and exploiting the role of context, and provide experiments on J-HMDB-21~\citep{jhuang2013towards} and UCF101-24~\citep{soomro2012ucf101}. We demonstrate that our approach increases the performance of three different video backbones, reaching state-of-the-art results on AVA 2.1 and AVA 2.2.}

\tit{Contributions} 
To sum up, our contributions are as follows:
\vspace{-.3cm}
\begin{itemize}
    \item \textcolor{black}{We propose a novel module for video action detection, which considers spatio-temporal relationships between actors and objects.}
    \vspace{-.3cm}
    \item \textcolor{black}{The proposed module is based on a spatio-temporal graph representation of the video, which is learned through self-attention operations. Further, multiple instances of the proposed module can be stacked together to obtain a greater temporal receptive field. The overall model is independent of the feature extraction stage and does not need end-to-end training to achieve state-of-the-art results.}
    \vspace{-.3cm}
    \item Extensive experiments on the challenging AVA dataset~\citep{gu2018ava} validate our approach and its components, demonstrating better performance with respect to other end-to-end models. In particular, our proposal achieves 29.8 and 31.8 mAP on AVA 2.1 and AVA 2.2, respectively, surpassing all previous approaches.
\end{itemize}

\section{Related work}
\label{sec:related}
\tit{Deep networks for video understanding}
CNNs are currently the state-of-the-art approach to extract spatio-temporal features for video processing and understanding~\citep{carreira2017quo,tran2015learning,varol2017long,xie2018rethinking}. \textcolor{black}{Most of the approaches for spatio-temporal feature extraction have employed either full 3D convolutional kernels, or a combination of 2D spatial kernels and 1D temporal filters~\citep{feichtenhofer2016spatiotemporal,qiu2017learning,tran2018closer}. Despite convolutions are capable of extracting temporal features to some extent, it is still a common practice to integrate both RGB and optical flow inputs in two separate streams~\citep{feichtenhofer2016convolutional,simonyan2014two,wang2016temporal} to capture appearance and motion respectively. Recently, some works have proposed architectural variations to improve the feature extraction capabilities of the network. \cite{li2019collaborative} have proposed to employ 2D convolutional kernels which slide over the three 2D projections of a spatio-temporal tensor; \cite{hussein2019timeception}, instead, have developed a multi-scale temporal convolution approach which uses different kernel sizes and dilation rates to better capture temporal dependencies. On the same line, \cite{feichtenhofer2019slowfast} has proposed a two-pathway network, with a low frame rate path that focuses on the extraction of spatial information and a high frame rate path that encodes motion.}

\tit{Detecting actions in space and time}
\textcolor{black}{Detecting actions is a fundamental step towards human behavior understanding in videos.
Towards this goal, two problems have been analyzed in the last few years: temporal action detection and spatio-temporal action detection. The former task aims to segment the temporal interval in which the action takes place~\citep{caba2015activitynet,sigurdsson2016hollywood,xu2017r}. The latter, instead, is intended to detect people in space and time and to classify their actions~\citep{gu2018ava,jhuang2013towards,soomro2012ucf101,soomro2014action}.
Seminal works in action detection and recognition have already investigated the role of context and that of modelling the interactions with objects~\citep{gupta2007objects,gupta2009observing,prest2012explicit,escorcia2013spatio} to improve recognition. Other approaches proposed to split the action localization task into spatial and temporal search~\citep{klaser2010human}.
}

\textcolor{black}{Recent approaches have tackled the spatio-temporal action detection task by exploiting human proposals coming from pre-trained image detectors~\citep{feichtenhofer2019slowfast,wu2019long} and replicating them in time to build straight spatio-temporal tubes; others have extended image detection architectures to infer more precise spatio-temporal tubelets~\citep{gkioxari2015finding,hou2017tube,kalogeiton2017action,saha2017amtnet}.
\cite{gu2018ava} proposed a baseline exploiting an I3D network encoding RGB and flow data separately, along with a Faster R-CNN~\citep{ren2015faster}, to jointly learn action proposals and labels. \cite{ulutan2020actor} suggested combining actor features with every spatio-temporal region in the scene to produce attention maps between the actor and the context. \cite{girdhar2019video}, instead, proposed a Transformer-style architecture~\citep{vaswani2017attention} to weight actors with features from the context around him. Finally, weakly-supervised approaches have also been proposed~\citep{escorcia2020guess}.
}

\tit{Graph-based representations}
\textcolor{black}{Graph-based representations have been used in action recognition~\citep{brendel2011learning,jain2016structural,wang2018videos,zhang2019structured} to model spatio-temporal relationships, although the use of graph learning and graph convolutional neural networks~\citep{defferrard2016convolutional,kipf2016semi,velivckovic2018graph} in video action detection is still under-investigated.
\cite{wang2018videos} proposed to model a video clip as a combination of the whole clip features and weighted proposal features, computed by a graph convolutional network based on similarities in the feature space and spatio-temporal distances between detections. \cite{zhang2019structured}, instead, defined the strength of a relation between two nodes in the graph as the inverse of the Euclidean distance between entities in the video.}

\section{Graph-based learning of spatio-temporal interactions}
\label{sec:approach}
\textcolor{black}{Given a video clip, the goal of our approach is to localize each actor and classify his actions. As actions performed in a clip depend on actor and object relationships through both space and time, we define a graph representation in which actor and object detections are treated as nodes, and edges hold relationships between them. Further, we link graphs from subsequent clips in time, to encode relations between clips belonging to the same longer video. We name our approach \ours, as an acronym of \textit{Spatio-Temporal Attention on Graph Entities}.}

\textcolor{black}{In this section, we first outline our graph-based representation for a single clip, describing the graph attention layer and the adjacency matrix we employ. In the remainder of the section, we will then extend this approach to handle a sequence of consecutive clips.}

\subsection{Graph-based clip representation}
\label{sec:graph}
\begin{figure*}[t]
    \centering
    \includegraphics[width=1\linewidth]{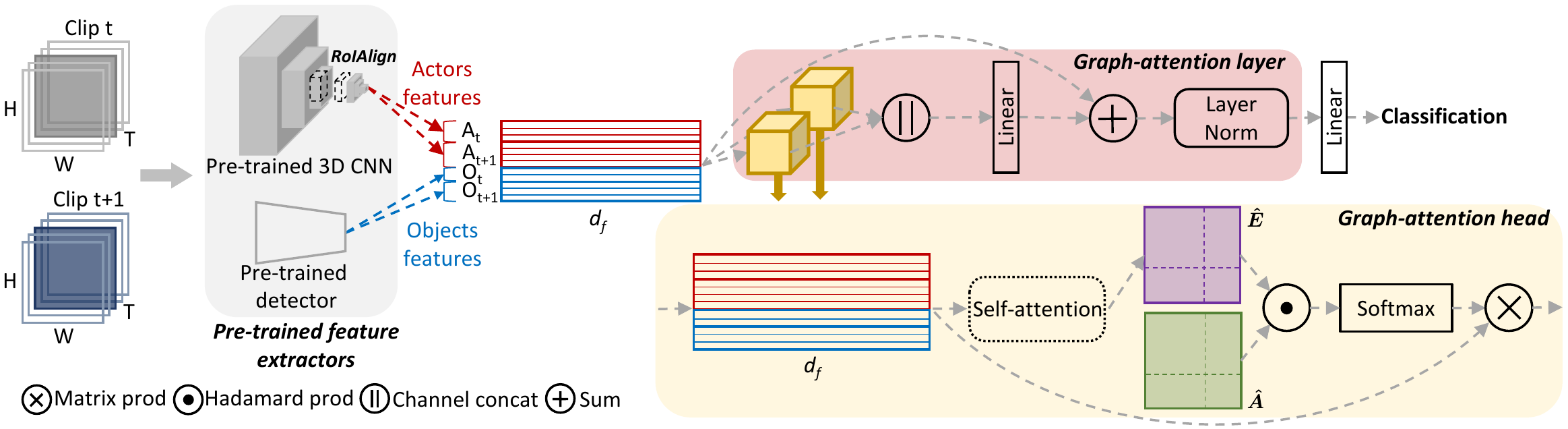}
    \caption{\textcolor{black}{Architecture of our high-level video understanding module. Given consecutive clips, the input of our model consists of actor and object features. It then applies a sequence of graph-attention layers (red-background box), each of them composed of a number of graph-attention heads (yellow-background box) applied in parallel. The figure depicts a single layer with two heads.}}
    \label{fig:model}
    \vspace{-.3cm}
\end{figure*}

We propose a graph-based representation of each clip, where nodes consist of actor and object features predicted by pre-trained detectors, as shown in the left side of Fig.~\ref{fig:model}. Denoting the number of actors and objects belonging to clip $t$ (\ie~centered in frame $t$) as $A_t$ and $O_t$ respectively, the total number of graph entities is $N_t=A_t+O_t$. Under this configuration, a clip can be represented as an $N_t \times d_f$ matrix, where $d_f$ is the node feature size. 

\textcolor{black}{Since actors can have meaningful relations both between them and with objects in the scene, we employ a fully-connected graph representation, in which all nodes are connected to the others, as the input of our network. Following the assumption that the closer an entity is to another, the higher the probability that they affect each other, a link between two entities in the graph is made stronger if they are spatially close. The graph configuration is therefore given by a dense $N_t \times N_t$ adjacency matrix $\bm{A}$, in which $\bm{A}_{ij}$ is defined as the proximity between entities $i$ and $j$, computed as follows:
\begin{equation}
\label{eq:adj_proximity}
\bm{A}_{ij} = e^{-\sqrt{(x_{ci}-x_{cj})^2 + (y_{ci}-y_{cj})^2}},
\end{equation}
}
where $x_{ci}$ and $y_{ci}$ are the center coordinates of entity $i$.
In the remainder of the paper, this single-clip adjacency matrix representation will be extended to a multi-clip adjacency matrix, allowing us to easily link graphs coming from subsequent clips of the same video.

\subsection{Spatial-aware graph attention}
\label{sec:gat}
\tit{Graph self-attention} 
\textcolor{black}{
Besides the introduced adjacency matrix, we adopt a graph attention module, inspired by~\cite{velivckovic2018graph}. The input of the model consists of $N_t$ node features which represent actors and objects, $\{\bm{f}_1,\bm{f}_2,...,\bm{f}_{N_t}\}$ with $\bm{f}_i \in \mathbb{R}^{d_f}$. First, the module applies a linear transformation to these features, in order to obtain a new representation of each entity $\{\bm{h}_1,\bm{h}_2,...,\bm{h}_{N_t}\}$, $\bm{h}_i \in \mathbb{R}^{d_h}$. Then a \textit{self-attention} operator $\mathcal{S}$ is applied to the nodes. In particular, the operator is defined as $\mathcal{S}:\mathbb{R}^{d_h}\times\mathbb{R}^{d_h}\rightarrow\mathbb{R}$, as follows:
\begin{equation}
\label{eq:attention1}
\bm{E}_{ij} = \mathcal{S}(\bm{h}_i,\bm{h}_j)
\end{equation}
with the scalar $\bm{E}_{ij}$ representing the \textit{importance} of entity $j$ with respect to entity $i$. Since we propose to represent a clip as a fully-connected graph, $\bm{E}_{ij}$ is computed for each pair of entities belonging to the same clip, avoiding the need for masking disconnected couples. Based on the original graph attention implementation~\citep{velivckovic2018graph}, $\mathcal{S}$ is implemented with a feedforward layer with $2 \times d_{h}$ parameters, followed by a LeakyReLU nonlinearity:
\begin{equation}
\label{eq:attention2}
\bm{E}_{ij} = \text{LeakyReLU}(\mathsf{FC}(\bm{h}_i \mathbin\Vert \bm{h}_j)),
\end{equation}
where $\mathbin\Vert$ indicates concatenation on the channel axis and $\mathsf{FC}$ is a linear layer. The resulting matrix, $\bm{E}$, will be a squared matrix with the same shape as the adjacency matrix. Separating it into its components, it can be rewritten as:
\begin{equation}
\label{eq:matrix_e}
\bm{E} = 
\begin{pmatrix} 
\bm{E}_{aa} & \bm{E}_{ao} \\
\bm{E}_{oa} & \bm{E}_{oo}
\end{pmatrix}
\end{equation}
where $\bm{E}_{aa}$ is the matrix of attentive weights between actors and actors, $\bm{E}_{ao}$ is the matrix of objects weights to actors, $\bm{E}_{oa}$ is the matrix of actors weights to objects and $\bm{E}_{oo}$ is the matrix of objects weights to objects.
}

\tit{Introducing spatial proximity}
The proposed self-attention module, when applied to a clip graph, computes the mutual influence of two entities in feature space, \ie~the influence of an entity on another based on their features. However, it does not consider mutual distances between entities.

\textcolor{black}{To introduce the prior given by the spatial proximity inside the clip, we condition the self-attention matrix $\bm{E}$ with the adjacency matrix $\bm{A}$, which contains the proximity between detections, by taking their Hadamard product, \ie:
\begin{equation}
\label{eq:hadamard}
\bm{D} = \bm{A} \odot \bm{E}.
\end{equation}
}
This operation allows us to strengthen the \textit{importance} of the features of an entity with respect to its neighbors and to weaken relations between entities that lie spatially far from each other. A row-wise softmax normalization is then applied to obtain an importance distribution over entities:
\textcolor{black}{
\begin{equation}
\label{eq:softmax_attention}
\bm{W}_{ij} = \frac{\exp(\bm{D}_{ij})}{\sum_{k=1}^{N_t} \exp(\bm{D}_{ik})}.
\end{equation}
The updated features computed by the module are a linear combination of the starting features $\{\bm{h}_1,\bm{h}_2,...,\bm{h}_{N_t}\}$ using $\{\bm{W}\}_{i,j}$ as coefficients. In particular, the self-attention module updates the initial features as follows:
\begin{equation}
\label{eq:weighted_sum}
\bm{h}'_{i} = \sigma\left(\sum_{j=1}^{N_t}\bm{W}_{ij}\bm{h}_{j}\right),
\end{equation}
where $\sigma$ is an ELU nonlinearity~\citep{clevert2016fast}.
}

\subsection{Temporal graph attention}
\label{sec:graphs_in_time}
In this section, we extend the proposed attention-based approach to jointly encode a batch of consecutive clips. Since different clips can have a different number of actors and objects, 
we devise a single adjacency matrix with as many rows and columns as the total number of entities in all clips of the batch. Besides allowing us to manage clips with a variable number of entities, this solution is suitable to link more graphs together and avoids padding. \textcolor{black}{When encoding a batch of consecutive clips, the size of the adjacency matrix will be $\sum_{t=1}^{b}N_t \times \sum_{t=1}^{b}N_t$, being $b$ the size of the batch of clips. }

\begin{figure}[t]
    \centering
    \includegraphics[width=0.7\linewidth]{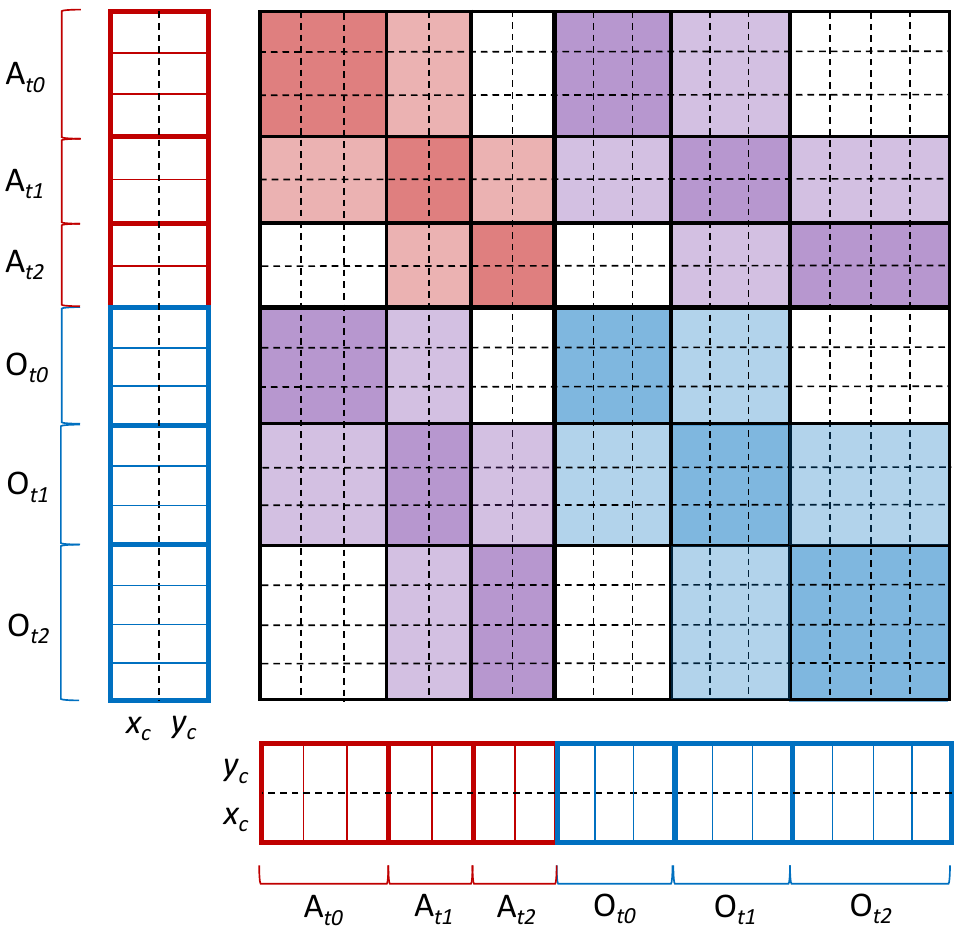}
    \caption{Adjacency matrix in a three clips per batch configuration, containing the spatial proximity between entities belonging to the same clip (dark-colored sub-matrices) and to consecutive clips (light-colored sub-matrices). White elements are zeros. Bounding box centers are indicated as $x_c$ and $y_c$.}
    \label{fig:adj_matrix}
    \vspace{-.3cm}
\end{figure}

An example is shown in Fig.~\ref{fig:adj_matrix}, for a three clips setting and a temporal receptive field of three consecutive clips. Here, dark red elements contain the proximity between actors of the same clip, dark blue elements contain the proximity between objects of the same clip, and dark violet elements contain the proximity between actors and objects of the same clip.
Entities belonging to subsequent clips can be linked by computing their boxes proximity (as in Eq.~\ref{eq:adj_proximity}), assuming that the temporal distance between clips is small enough to ensure the consistency of the scene. The light-colored elements of the adjacency matrix in Fig.~\ref{fig:adj_matrix} contain the proximity between actors (light red), objects (light blue), and actors/objects (light violet) belonging to two consecutive clips. The temporal receptive field of a single attentive layer can be potentially increased by adding the proximity of temporally distant entities in the adjacency matrix. 

\tit{Self-attention over time}
We extend the self-attentive operations to compute the \textit{importance} of an entity with respect to all the other entities in the batch. \textcolor{black}{In our implementation, the \textit{self-attention} module computes attention weights for each pair of entity features, without any masking. For a three clips per batch setting, the complete attention weights matrix $\hat{\bm{E}}$ looks like the following:
\begin{equation}
\label{eq:matrix_E}
\hat{\bm{E}} = 
\left( \begin{array}{@{}c|c@{}}
  \begin{matrix}
      \hat{\bm{E}}_{aa}^{t_0,t_0} & \hat{\bm{E}}_{aa}^{t_0,t_1} & \hat{\bm{E}}_{aa}^{t_0,t_2} \\
      \hat{\bm{E}}_{aa}^{t_1,t_0} & \hat{\bm{E}}_{aa}^{t_1,t_1} & \hat{\bm{E}}_{aa}^{t_1,t_2} \\
      \hat{\bm{E}}_{aa}^{t_2,t_0} & \hat{\bm{E}}_{aa}^{t_2,t_1} & \hat{\bm{E}}_{aa}^{t_2,t_2}
  \end{matrix} 
      & \begin{matrix}
      \hat{\bm{E}}_{ao}^{t_0,t_0} & \hat{\bm{E}}_{ao}^{t_0,t_1} & \hat{\bm{E}}_{ao}^{t_0,t_2} \\
      \hat{\bm{E}}_{ao}^{t_1,t_0} & \hat{\bm{E}}_{ao}^{t_1,t_1} & \hat{\bm{E}}_{ao}^{t_1,t_2} \\
      \hat{\bm{E}}_{ao}^{t_2,t_0} & \hat{\bm{E}}_{ao}^{t_2,t_1} & \hat{\bm{E}}_{ao}^{t_2,t_2}
  \end{matrix}  \\
  \cmidrule[0.4pt]{1-2}
  \begin{matrix}
      \hat{\bm{E}}_{oa}^{t_0,t_0} & \hat{\bm{E}}_{oa}^{t_0,t_1} & \hat{\bm{E}}_{oa}^{t_0,t_2} \\
      \hat{\bm{E}}_{oa}^{t_1,t_0} & \hat{\bm{E}}_{oa}^{t_1,t_1} & \hat{\bm{E}}_{oa}^{t_1,t_2} \\
      \hat{\bm{E}}_{oa}^{t_2,t_0} & \hat{\bm{E}}_{oa}^{t_2,t_1} & \hat{\bm{E}}_{oa}^{t_2,t_2}
  \end{matrix}  & 
  \begin{matrix}
      \hat{\bm{E}}_{oo}^{t_0,t_0} & \hat{\bm{E}}_{oo}^{t_0,t_1} & \hat{\bm{E}}_{oo}^{t_0,t_2} \\
      \hat{\bm{E}}_{oo}^{t_1,t_0} & \hat{\bm{E}}_{oo}^{t_1,t_1} & \hat{\bm{E}}_{oo}^{t_1,t_2} \\
      \hat{\bm{E}}_{oo}^{t_2,t_0} & \hat{\bm{E}}_{oo}^{t_2,t_1} & \hat{\bm{E}}_{oo}^{t_2,t_2}
  \end{matrix} \\
\end{array} \right)
\end{equation}
where $\hat{\bm{E}}_{aa}^{t,t'}$ is the weights matrix of actors belonging to clip $t$ to actors belonging to clip $t'$, $\hat{\bm{E}}_{ao}^{t,t'}$ is the weights matrix of objects belonging to clip $t$ to actors belonging to clip $t'$, and so on.
}

\noindent \textcolor{black}{Given the new adjacency matrix, $\hat{\bm{A}}$, the Hadamard product:
\begin{equation}
\label{eq:hadamard2}
\hat{\bm{D}} = \hat{\bm{A}} \odot \hat{\bm{E}}
\end{equation}}
is in charge of strengthening or weakening (based on the spatial distance) weights between entities belonging to the same timestamp or sufficiently close in time, and to zero weights between temporally distant entities.
Finally, the linear combination of Eq.~\ref{eq:weighted_sum} replaces features of an entity with a weighted sum of features directly connected to it in the graph: these features come now from entities belonging to the same clip and to temporally close clips. As our approach works at the feature level, increasing the temporal receptive field of the module does not dramatically increase the demand of computational resources.

\tit{Multi-head multi-layer approach}
A single graph-attention head (yellow-background box of Fig.~\ref{fig:model}) performs the aforementioned operations in our model. A graph-attention layer (red-background box of Fig.~\ref{fig:model}) concatenates the output of different heads and applies a linear layer, a residual connection, and a layer normalization~\citep{ba2016layer}.
As it will be analyzed in Sec.~\ref{sub:ablation}, a grid search on the number of parallel heads and subsequent layers allows us to obtain the best performance.

It is worth noting that the number of graph attention layers affects the temporal receptive field of the overall sequence of layers. Considering a receptive field of three (corresponding to a graph where entities are directly connected only with other entities of the same clip and to entities from the previous and following clips), each layer after the first one increases the overall temporal receptive field by two. 
In a two layers setting, for instance, the second graph attention layer will compute the features of a clip as a weighted sum of its two neighbors, but features from those have already been affected by features of other clips in the first graph attention layer.

\section{Experimental Results}
\label{sec:experiments}
\textcolor{black}{
In this section, we introduce the experimental setting and report the implementation and training details. We then provide quantitative and qualitative evaluations, as well as computational analysis.
}

\tit{Datasets and metrics}
\label{sub:dataset}
\textcolor{black}{
We evaluate our model on versions 2.1 and 2.2 of the challenging AVA dataset~\citep{gu2018ava}, which contains annotations to localize people both in space and time and to predict their actions.} It consists of 235 training and 64 validation videos, each 15 minutes long. The temporal granularity of annotations is 1 second, leading to 211k training and 57k validation clips centered in the annotated keyframes. Each actor is involved in one or more of 80 atomic action classes.
One of the main challenges of AVA concerns its long-tail property: tens of thousands of samples are available for some classes while only a few dozen for others. 

The performance of a model on AVA is measured by the keyframe-level mean Average Precision (mAP) with a 50\% IoU threshold. Following the protocol suggested by the dataset authors and adopted in prior works, we train our architecture on all the 80 classes and evaluate its performance only on the 60 classes containing at least 25 validation examples.

\textcolor{black}{We also report performances in terms of frame-mAP with the same 50\% IoU threshold on two additional benchmarks, namely J-HMDB-21~\citep{jhuang2013towards} and UCF101-24~\citep{soomro2012ucf101}. These two datasets are relatively smaller than AVA, and provide a single label per video. On average, they also have fewer interactions between entities.}

\tit{\textcolor{black}{People and object detectors}}
\label{sub:detection}
\textcolor{black}{When experimenting on AVA, we use a Faster R-CNN~\citep{ren2015faster} with a ResNeXt-101-FPN~\citep{he2016deep,lin2017feature,xie2017aggregated} as people detector, applied on keyframes. The detection network is pre-trained on COCO~\citep{lin2014microsoft} and fine-tuned on AVA~\citep{gu2018ava} people boxes.} The detector is the same used by~\cite{feichtenhofer2019slowfast} and~\cite{wu2019long}, and reaches 93.9 AP@50
on the AVA validation set. Following previous works~\citep{feichtenhofer2019slowfast,gu2018ava}, actor features are then obtained from a 3D CNN backbone (which is discussed in the next section) by replicating boxes in time to obtain a 3D RoI and applying RoIAlign~\citep{he2017mask}. During training, we employ ground-truth people regions. Objects features, instead, are extracted from a Faster R-CNN detector pre-trained on Visual Genome~\citep{krishna2017visual}, and have a dimensionality of 2048.


\tit{\textcolor{black}{Video Backbones}}
\textcolor{black}{In all experiments, we employ a pre-trained actor backbone which is kept fixed during training. Freezing the backbone allows us to increase the batch size and to explore longer temporal relations between consecutive clips. The pre-trained backbones take raw clips as input and output features for each actor.} \textcolor{black}{When considering the AVA dataset, all video-level backbones are trained on the Kinetics dataset~\citep{kay2017kinetics} and fine-tuned on the AVA dataset~\citep{gu2018ava} before applying our module.} 

\textcolor{black}{On AVA, we consider three backbones for extracting actor features, namely I3D~\citep{carreira2017quo}, R101-I3D-NL~\citep{wu2019long} and SlowFast-NL~\citep{feichtenhofer2019slowfast}.} The I3D~\citep{carreira2017quo} backbone is pre-trained on ImageNet~\citep{deng2009imagenet} before being ``inflated'', and then trained on Kinetics-400. RoIAlign is applied after the \textit{Mixed\_4f} layer and we fine-tune only the last layers (from the \textit{Mixed\_5a} layer to the final linear classifier) on AVA for 10 epochs. The R101-I3D-NL~\citep{wu2019long} and the SlowFast-NL~\citep{feichtenhofer2019slowfast} backbones are pre-trained on Kinetics-400 or Kinetics-600 (R101-I3D-NL on ImageNet, too), and fine-tuned end-to-end on the AVA dataset. 

For the I3D backbone, we use ground-truth boxes and predicted boxes with any score during features extraction, assigning labels of a ground-truth box to a predicted box if their IoU is 0.5 or more. We use predicted boxes with score at least 0.7 for the evaluation. Following the authors implementation~\citep{feichtenhofer2019slowfast,wu2019long}, for the R101-I3D-NL and SlowFast-NL backbones we use ground-truth boxes and predicted boxes with score at least 0.9 during features extraction, assigning labels of a ground-truth box to a predicted box if their IoU is 0.9 or more. We use predicted boxes with score at least 0.8 for the evaluation. 

\textcolor{black}{Features are always extracted from the last layer of the backbone before classification, after averaging in space and time dimensions: feature size, therefore, is 1024, 2048, and 2304 for I3D, R101-I3D-NL, and SlowFast-NL respectively. As additional features, we also add bounding boxes height, width and center coordinates to actors and objects, as we found it to be beneficial in preliminary experiments. 
A linear layer is employed to transform actor or object features to a common dimensionality $d_f$, making their concatenation feasible. In all experiments, when concatenating actor and object features we apply the linear layer to the feature vector with the largest dimensionality, leaving the other unchanged.}

\tit{Implementation and training details}
Each graph attention head consists of two fully-connected layers. The first one reduces the feature size depending on the number of heads used in that layer: with $n_h$ heads, the output feature size is set to $\left \lfloor{d_f / n_h}\right \rfloor $. The second linear layer, instead, computes attention weights (Eq.~\ref{eq:attention2}). The outputs of different attention heads are then concatenated, and a fully connected layer followed by a residual block and a layer normalization block is applied (Fig.~\ref{fig:model}). \textcolor{black}{Each graph attention head is followed by a dropout with keep probability 0.5, and the alpha parameter of the LeakyReLU in Eq.~\ref{eq:attention2} is set to 0.2. After a sequence of layers of the proposed module, one last linear layer is employed to compute per-class probabilities, and a sigmoid cross-entropy loss (for AVA) or a softmax cross-entropy loss (for the other benchmarks) is applied.
In our experiments, we adopt a temporal receptive field of three, connecting entities of a clip with those belonging to the same, the previous and the following clip. Table~\ref{tab:architecture} lists the learnable blocks of our architecture in a 4-heads/1-layer setting. It is worthwhile to mention that our graph-attention block is trained without any data augmentation, while end-to-end approaches typically require random flipping, scaling, and cropping.}

\begin{table}[t]
\small
\scriptsize
\centering
\begin{tabular}{lcccccc}
\toprule 
Stage & & Module & & Input size & & Output size \\
\midrule
$Input$ & & & & $N_t \times 1024$ & &  \\
\midrule
$GAL_1$     & & $\mathsf{FC}_{11}$ & & $\left[N_t \times 1024\right] \times 4$           & & $\left[N_t \times 256\right] \times 4$ \\
            & & $\mathsf{FC}_{12}$ & & $\left[N_t \times N_t \times 512\right] \times 4$ & & $\left[N_t \times N_t\right] \times 4$ \\
            & & $\mathsf{FC}_{13}$ & & $N_t \times 1024$                                 & & $N_t \times 1024$ \\
            & & $\mathsf{LNorm}$   & & $N_t \times 1024$                                 & & $N_t \times 1024$ \\
\midrule 
            & & $\mathsf{FC}_{3}$ & & $A_t \times 1024$ & & $A_t \times classes$\\
\bottomrule
\end{tabular}
\caption{\textcolor{black}{Learnable blocks of \ours, in a 4-heads/1-layer configuration, when using I3D features. $GAL_i$ indicates the $i$-th graph-attention-layer, which consists of 4 attention heads (each with 2 fully-connected layers), a linear layer, and a LayerNorm. $N_t$ is the number of entities (actors and objects), $A_t$ is the number of actors.}}
\label{tab:architecture}
\vspace{-.3cm}
\end{table}

During training, we use a batch size of 6. Adam optimizer~\citep{kingma2014adam} is adopted in all our experiments, with a learning rate of $6.25 \times 10^{-5}$ when using I3D features and $10^{-5}$ for R101-I3D-NL and SlowFast-NL features. The learning rate is decreased by a factor of 10 when the validation mAP does not increase for ten consecutive epochs. Early-stopping is also applied when the validation mAP does not increase for five consecutive epochs. All the experiments are performed on a single NVIDIA V100 GPU; on average, a single experiment takes less than a day to converge.

\begin{table}[t]
\scriptsize
\centering
\begin{tabular}{lcc}
\toprule 
Model & Pretraining & mAP@50  \\
\midrule
AVA~\citep{gu2018ava}                                    & Kinetics-400 & 15.6 \\
ACRN~\citep{sun2018actor}                                & Kinetics-400 & 17.4 \\
STEP~\citep{yang2019step}                                & Kinetics-400 & 18.6 \\
Better baseline~\citep{girdhar2018better}                & Kinetics-600 & 21.9 \\
SMAD~\citep{zhang2019structured}                         & Kinetics-400 & 22.2 \\
RTPR~\citep{Li_2018_ECCV}                                & - & 22.3 \\
ACAM~\citep{ulutan2020actor}                             & Kinetics-400 & 24.4 \\
VATX~\citep{girdhar2019video}                            & Kinetics-400 & 24.9 \\
SlowFast~\citep{feichtenhofer2019slowfast}               & Kinetics-400 & 26.3 \\
LFB (R101-I3D-NL)~\citep{wu2019long}                     & Kinetics-400 & 26.8 \\
\midrule
I3D~\citep{carreira2017quo}                              & Kinetics-400 & 19.7 \\
\textbf{\ours (I3D)}             & Kinetics-400 & \textbf{23.0} \\
R101-I3D-NL~\citep{wu2019long}                           & Kinetics-400 & 23.9\\
\textbf{\ours (R101-I3D-NL)}          & Kinetics-400 & \textbf{26.3} \\
SlowFast-NL,8$\times$8~\citep{feichtenhofer2019slowfast}             & Kinetics-600 & 28.2 \\
\textbf{\ours (SlowFast-NL,8$\times$8)}      & Kinetics-600 & \textbf{29.8} \\
\bottomrule
\end{tabular}
\caption{\textcolor{black}{Comparison with previous approaches on AVA 2.1 validation set, in terms of mean Average Precision.}}
\label{tab:map}
\end{table}

\subsection{Main quantitative results}
In the following experiments, we show that our proposed module can improve video action detection performance by a significant margin, reaching state-of-the-art results. If not specified otherwise, we employ a 2-layers 4-heads setting when using the I3D backbone and when testing on J-HMDB-21 and UCF101-24, and a 2-layers 2-heads setting when using the R101-I3D-NL and the SlowFast-NL backbones.

\tit{Results on AVA 2.1}
Table~\ref{tab:map} shows the mean Average Precision with 50\% IoU threshold on AVA 2.1 for our method, considering the three backbones, and for a number of competitors. All the experiments refer to a single-crop validation accuracy (no multi-scale and horizontal flipping are adopted for testing) and for a single model (\ie, without using ensemble methods).

\begin{table}[t]
\scriptsize
\centering
\begin{tabular}{lccc}
\toprule 
Model & Pretraining & \begin{tabular}{@{}c@{}}val \\ mAP\end{tabular} & \begin{tabular}{@{}c@{}}test \\ mAP\end{tabular} \\
\midrule
SlowFast-NL,8$\times$8~\citep{feichtenhofer2019slowfast}*                    & Kinetics-600 & 29.1  & -\\
\textbf{\ours (SlowFast-NL,8$\times$8)}     & Kinetics-600 & \textbf{30.0} & \textbf{29.6} \\
SlowFast-NL,16$\times$8~\citep{feichtenhofer2019slowfast}*     & Kinetics-600 & 29.4 & -\\
\textbf{\ours (SlowFast-NL,16$\times$8)}   & Kinetics-600 & \textbf{30.3} & \textbf{29.9} \\
\midrule
\textbf{\ours (SlowFast-NL,16$\times$8,8$\times$8)}   & Kinetics-600 & \textbf{31.8} & \textbf{31.6} \\
\bottomrule
\end{tabular}
\caption{\textcolor{black}{Comparison with previous approaches on 
AVA 2.2 validation and test sets, using different backbones.} Numbers marked with ``*" are obtained from models released in the official SlowFast~\citep{feichtenhofer2019slowfast} repository. In~\citep{feichtenhofer2019slowfast}, the reported performances are 29.0 and 29.8 for SlowFast-NL,8x8 and SlowFast-NL,16x8 respectively.}
\label{tab:map2}
\vspace{-.2cm}
\end{table}

When applying our approach, we observe a relative improvement of more than 16\% on the I3D backbone (19.7 $\rightarrow$ 23.0), of about 10\% for the R101-I3D-NL backbone (23.9 $\rightarrow$ 26.3) and of almost 6\% for the SlowFast-NL backbone (28.2 $\rightarrow$ 29.8). Noticeably, the presence of non-local operations~\citep{wang2018non} in these last two backbones does not prevent our model from improving performance, underlying that these two techniques are complementary. Also, the results obtained using the I3D backbone are superior to many approaches that employ the same backbone and train end-to-end. The adoption of a long-term feature bank in~\cite{wu2019long} brings slightly better performance (26.8) compared to our solution (26.3) using the same R101-I3D-NL backbone. It is worth noting, although, that~\cite{wu2019long} uses two instances of the backbone, one to compute long-term and another to compute short-term features, both fine-tuned end-to-end. Our model, instead, uses only one backbone instance, which is also kept fixed during training. Single-crop validation mAP obtained with the SlowFast-NL backbone (\textbf{29.8 mAP}) represents a new state of the art for the AVA v2.1 dataset. 

\begin{table}[t]
\setlength{\tabcolsep}{5pt}
\scriptsize
\centering
\begin{tabular}{lccc}
\toprule 
Model & Pose & Person-Person & Person-Object \\
\midrule
I3D~\citep{carreira2017quo} & 37.4 & 20.4 & 12.2 \\
\textbf{\ours(I3D)} & \textbf{40.4} & \textbf{23.5} & \textbf{15.7} \\
\midrule
R101-I3D-NL~\citep{wu2019long} & 41.4 & 26.5 & 15.5 \\
\textbf{\ours(R101-I3D-NL)} & \textbf{43.4} & \textbf{29.0} & \textbf{18.1} \\
\midrule
SlowFast-NL~\citep{feichtenhofer2019slowfast} & 48.3 & 28.8 & 19.8 \\
\textbf{\ours(SlowFast-NL)} & \textbf{50.3} & \textbf{31.4} & \textbf{20.8}
\\
\bottomrule
\end{tabular}
\caption{\textcolor{black}{Mean Average Precision on different AVA 2.1 class groups (actions involving person-pose, person-person and person-object interactions), when using the I3D backbone.}}
\label{tab:action_type}
\end{table}

\begin{table}[t]
\scriptsize
\centering\color{black}
\begin{tabular}{lcc}
\toprule
Model$\downarrow$~~Dataset$\rightarrow$ & J-HMDB-21 & UCF101-24 \\
\midrule
Action tubes~\citep{gkioxari2015finding} & 27.0 & - \\
Action tubes*~\citep{gkioxari2015finding} & 28.6 & - \\
\ours (Action tubes*) & \textbf{29.6} & - \\
\midrule
AMTnet~\citep{saha2017amtnet} & 45.0 & - \\
AMTnet*~\citep{saha2017amtnet} & 47.0 & 67.7 \\
\textbf{\ours (AMTnet*)} & \textbf{48.1} & \textbf{69.1} \\
\bottomrule
\end{tabular}
\caption{Experimental results on J-HMDB-21 and UCF101-24, using the detection backbones presented in~\cite{gkioxari2015finding} and~\cite{saha2017amtnet}. For both backbones, the three rows report the mAP presented in the original paper, the mAP we obtained training a linear classifier on top of pre-extracted features (marked by *) and the mAP of STAGE on the same features, respectively.
}
\label{tab:additional_datasets}
\vspace{-.3cm}
\end{table}

\tit{Results on AVA 2.2}
Table~\ref{tab:map2} reports the performance of our approach on the more recent AVA v2.2. Here, the test mAP has been computed using the official AVA evaluation server, after training on both train and val splits, following the common practice in literature. As it can be observed, adopting \ours on top of SlowFast-NL,8x8 is better than doubling the number of input frames to the backbone (\ie, using SlowFast-NL,16x8). \textcolor{black}{This underlines that modelling high-level entities is at least as important as extracting better spatio-temporal features.}

\textcolor{black}{Finally, leveraging the fact that \ours is backbone independent, we train it using both SlowFast-NL,8x8 and SlowFast-NL,16x8 features, which are averaged before being forwarded through \ours. This model achieves single-crop \textbf{31.8 mAP}, a new state of the art for AVA v2.2.} 

\begin{figure}[t]
\centering
\scriptsize
\setlength{\tabcolsep}{.15em}
\scriptsize
\begin{tabular}{c}
\includegraphics[width=0.75\linewidth]{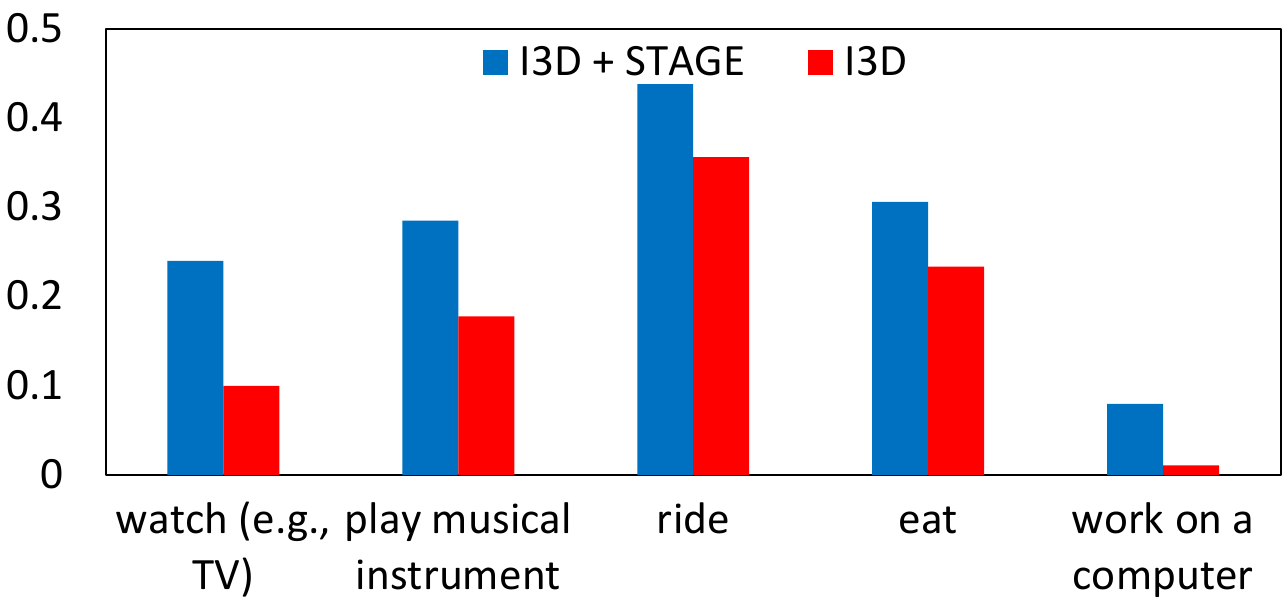} \\
\includegraphics[width=0.75\linewidth]{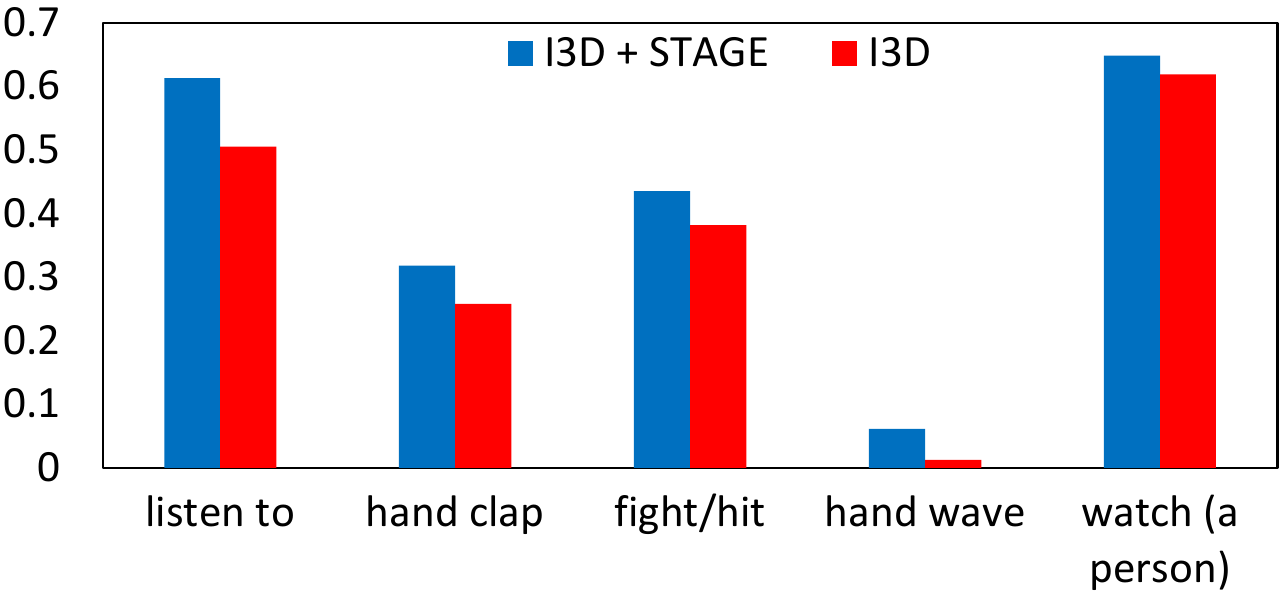} \\
\end{tabular}
\caption{Per-class Average Precision of an I3D backbone with and without our module. We report the five classes with the highest absolute gain among person-object interaction classes (top) and person-person interaction classes (bottom).}
\label{fig:histogram}
\vspace{-.3cm}
\end{figure}

\tit{Per-class analysis}
In Table~\ref{tab:action_type}, we show the performances of our approach on different AVA class groups, \ie~actions involving person-pose (13 classes), person-person interactions (15 classes) and person-object interactions (32 classes). As it can be seen, our model shows a higher improvement for actions that involve an interaction between entities (which are also the majority in AVA). We also note that the recognition of pose classes (like \textit{dance} or \textit{martial art}) benefits from interactions and elements from the context.
Finally, in Fig.~\ref{fig:histogram}, we show the five person-object interaction classes (top) and five person-person interaction classes (bottom) with the top AP gain, when considering the I3D backbone. Classes with the highest absolute gain are \textit{watch (e.g., TV)} (+14.0 AP), \textit{listen to (a person)} (+10.7 AP), \textit{play musical instrument} (+ 10.7 AP), all involving interactions with other objects or actors.


\tit{\textcolor{black}{Results on J-HMDB-21 and UCF101-24}}
\textcolor{black}{We also experimented the capabilities of STAGE on two additional benchmarks, namely J-HMDB-21~\citep{jhuang2013towards} and UCF101-24~\citep{soomro2012ucf101}, in comparison with the models presented in~\cite{gkioxari2015finding} and in~\cite{saha2017amtnet}. For fairness of evaluation, we adopt STAGE on top of the actor detection backbones presented in~\cite{gkioxari2015finding} and in~\cite{saha2017amtnet}. Hence, only actor boxes are considered, and no objects. J-HMDB-21 mAP is averaged over the three splits, while UCF101-24 mAP refers to the first split of the dataset, following the standard practice in literature.}
\textcolor{black}{Results are reported in Table~\ref{tab:additional_datasets}, where, for each dataset, we also report the frame-mAP obtained training a linear classifier on top of pre-extracted features (marked with *), and the frame-mAP obtained through STAGE applied on the same features. Only RGB features are considered, without exploiting optical flow. As it can be seen, in both settings applying STAGE leads to a performance improvement of around 1 mAP point.
}

\subsection{Ablation study}
\label{sub:ablation}
To validate the importance of the design choices made in our graph-attention module, we run several ablation experiments. We explore different combinations of heads and layers, remove key components in the architecture, and modify the graph structure to use existing techniques in place of our choices. \textcolor{black}{In all the following experiments we employ AVA v2.1.}

\tit{Varying the number of heads and layers}
Table~\ref{tab:abl1} shows the effect of varying the number of graph-attention heads and layers when using \ours with features from the three adopted backbones. As it can be noticed, stacking multiple layers together brings better performance: each layer after the first increases the temporal receptive field, generating indirect edges between temporally distant nodes. The best configuration is obtained when using a 4-heads/2-layers setting for I3D features and a 2-heads/2-layers setting for R101-I3D-NL and SlowFast-NL. These configurations are also used in the following ablation experiments.

\begin{table*}[t]
\scriptsize
\centering
\begin{tabular}{lccccccccccccc}
\toprule 
Layers$\rightarrow$ & & & & 1 & & & & 2 & & & & 3 & \\
\cmidrule{4-6}\cmidrule{8-10}\cmidrule{12-14}
Heads$\rightarrow$ & & & 2 & 4 & 8 & & 2 & 4 & 8 & & 2 & 4 & 8 \\
\midrule
\ours(I3D) & & & 21.2 & 21.7 & 22.0 & & 21.7 & \textbf{23.0} & 22.8 & & 21.7 & 22.7 & 21.9 \\
\ours(R101-I3D-NL) & & & 26.2 & 26.1 & \textbf{26.3} & & \textbf{26.3} & 26.1 & 25.8 & & \textbf{26.3} & 26.0 & 25.6 \\
\ours(SlowFast-NL) & & & 29.7 & 29.2 & 29.6 & & \textbf{29.8} & 29.3 & 29.6 & & 29.6 & 29.3 & 29.5 \\
\bottomrule
\end{tabular}
\caption{Validation mAP obtained considering different combinations of graph-attention heads and layers.}
\label{tab:abl1}
\vspace{-.3cm}
\end{table*}

\begin{table}[t]
\scriptsize
\centering\color{black}
\begin{tabular}{lccc}
\toprule 
Head$\downarrow$~~Backbone$\rightarrow$ & I3D & R101-I3D-NL & SlowFast-NL \\
\midrule
1L STO & 20.2 & 24.7 & 28.5 \\
2L STO & 20.4 & 25.0 & 28.7 \\
STO (\ours) & 20.5 & 25.5 & 29.5 \\
\textbf{\ours} & \textbf{23.0} & \textbf{26.3} & \textbf{29.8} \\ 
\bottomrule
\end{tabular}
\caption{Comparison with the STO baseline.}
\label{tab:STO_baseline}
\end{table}

\begin{table}[t]
\scriptsize
\centering
\begin{tabular}[t]{lccc}
\toprule 
Model$\downarrow$~~Backbone$\rightarrow$ & I3D & R101-I3D-NL & SlowFast-NL\\
\midrule
Original Backbone & 19.7 & 23.9 & 28.2 \\
\textbf{\ours} & \textbf{23.0} & \textbf{26.3} & \textbf{29.8} \\
\midrule
\ours w/o boxes proximity & 21.5 & 25.8 & 29.6\\
\ours w/o temporal connections & 22.1 & 25.7 & 29.4\\
\ours w/o actors-actors interactions & 22.1 & 26.1 & 28.9\\
\ours w/o object-object interactions & 22.3 & 25.6 & \textbf{29.8} \\
\midrule
\ours w/ Transformer attention & 21.6 & 25.6 & 29.6 \\
\ours w/ nodes Euclidean distance & 21.4 & \textbf{26.3} & 29.5\\
\bottomrule
\end{tabular}
\caption{Validation mAP obtained removing some key components in our model or replacing them with other existing techniques.}
\label{tab:abl2}
\vspace{-.3cm}
\end{table}

\tit{\textcolor{black}{Comparison with the STO baseline}}
\textcolor{black}{In Table~\ref{tab:STO_baseline} we compare the STAGE module with the STO operator~\citep{wu2019long} applied on top of the considered backbones. The STO baseline consists of a short-term operator that updates actor features on the basis of other actors from the same clip, using one or more non-local blocks~\citep{wang2018non}. STO lacks the graph structure, the object detections, and the temporal interactions, leading to worse performances compared to STAGE (Table~\ref{tab:STO_baseline} first two rows, corresponding to one-layer and two-layers STO, respectively). In the third row of Table~\ref{tab:STO_baseline}, instead, we report the performance of the STAGE module when replacing Eq.~\ref{eq:attention2} with the non-local-based attention, and removing temporal interactions. The original STAGE design demonstrates higher mAP in this case, too.
}

\tit{Removing key components}
In Table~\ref{tab:abl2} we report the validation mAP obtained when removing key components. Performances drop when removing the spatial prior between detections. Moreover, when the temporal links between consecutive clips are removed and only edges between nodes of the same clip are kept, we observe a reduction in mAP. To evaluate the design of the graph structure, we first remove actor-actor interactions to quantify the role of objects: in this setting, actor features are updated with a weighted sum of object features. As it can be seen, this leads to better performances compared to those of graph-free backbones. One can question if object-object interactions are useful: when removing them from the graph, performances drop for both I3D and R101-I3D-NL backbones. Our insight is that some recurring combinations of objects can be useful at prediction time: a closed door in clip $t$, for instance, related to the same open door in clip $t+1$ could help to recognize the \textit{open} action.

\tit{Attention and adjacency alternatives}
In the last two rows of Table~\ref{tab:abl2} we show results obtained by replacing our attention mechanism and our adjacency matrix design with other proposals. We investigate the use of dot-product attention, by replacing the weights of Eq.~\ref{eq:attention2} with weights computed through a Transformer-like self-attention~\citep{vaswani2017attention}, as follows:
\textcolor{black}{
\begin{equation}
\label{eq:attention_transformer}
\bm{E} = \frac{\bm{Q}\bm{K}^{T}}{\sqrt{d_k}} \bm{V},
\end{equation}
where $\bm{Q}$, $\bm{K}$ and $\bm{V}$ come from three linear projections of input features. In this setting, as it can be noticed, we again observe a significant drop in performance. 
}

Taking inspiration from~\cite{zhang2019structured}, we also replace the Euclidean distance between bounding box coordinates with the Euclidean distance between bounding box features in the adjacency matrix. We found this choice to lower the performance on both I3D and SlowFast-NL backbones. In this setting, Eq.~\ref{eq:adj_proximity} is replaced by:
\textcolor{black}{
\begin{equation}
\label{eq:adj_proximity_nodes}
\bm{A}_{ij} = \frac{1}{||\bm{h}_i - \bm{h}_j||_2}.
\end{equation}
}

It shall be noted that all the aforementioned ablations do not change the number of parameters in the model (except for the Transformer attention experiment, where each attention head uses three linear layers instead of two), thus confirming the effectiveness of our approach. 
We finally note that \ours with SlowFast-NL,8$\times$8 backbone reaches 36.5 validation mAP on AVA 2.1 when tested with ground-truth actor boxes, suggesting that a stronger person detector could significantly boost performances.

\subsection{Computational analysis}
\textcolor{black}{Our module can reach state-of-the-art results without requiring end-to-end training of the backbone. This has an impact on the computational requirements of \ours at training time, since the convolutional backbone incorporates most of the model complexity. Table~\ref{tab:gpus} shows a comparison with different approaches employing end-to-end training in terms of training time and resource requirements.} For each approach, we report the number of GPUs used during training, the batch size per GPU, the number of epochs, and the overall training time. The comparison is based on the implementation details reported in the original papers and refers to a training on the AVA~\citep{gu2018ava} dataset. Our module requires a single GPU for training when pre-extracting backbone features, and less than a day to converge.

\textcolor{black}{Finally, in Table~\ref{tab:FLOPs_Params}, we report the additional FLOPs and trained parameters introduced by STAGE during inference. Please note that both the number of floating-point operations and the number of trained parameters depend on the dimensionality of the features produced by each backbone. As the amount of FLOPs also depends on the number of detections predicted on each clip, we consider a clip with a number of actor and object detections equal to the average number of actors and objects in all AVA training clips, \ie~4 actors and 25 objects.}

\begin{table}[t]
\setlength{\tabcolsep}{3.5pt}
\small
\scriptsize
\centering
\begin{tabular}{lccccc}
\toprule 
Model & \# GPUs & Clips/GPU & Epochs & Training time  \\
\midrule
ACRN~\citep{sun2018actor}                            & 11 & 1 & 63 &  - \\
Better baseline~\citep{girdhar2018better}            & 11 & 3 & 78 &  - \\
SMAD~\citep{zhang2019structured}                     & 8  & 2 & 11 &  - \\
VATX~\citep{girdhar2019video}                        & 10  & 3 & 71 & $\sim$ 7 days \\
SlowFast~\citep{feichtenhofer2019slowfast}           & 128 & - & 68 & - \\
LFB~\citep{wu2019long}                               & $8\times2$  & 2 & $10\times2$ & $\sim$ $2\times2$ days \\
\midrule
\textbf{\ours}    & 1  & 6 & 20 & $<$ 1 day \\
\bottomrule
\end{tabular}
\caption{\textcolor{black}{Training times and computational requirements of \ours and existing approaches involving an end-to-end finetuning of the 3D backbone. The LFB model uses two instances of the backbone, thus has twice the complexity of its base model.}}
\label{tab:gpus}
\end{table}

\begin{table}[t]
\setlength{\tabcolsep}{3.5pt}
\small
\scriptsize
\centering\color{black}
\begin{tabular}{lccccc}
\toprule 
Model & GFLOPs & Parameters  \\
\midrule
I3D~\citep{carreira2017quo} & 108 & 12M \\
+\textbf{\ours} & +0.11 & +6.4M \\
R101-I3D-NL~\citep{wu2019long} & 359 & 54.3M \\
+\textbf{\ours} & +0.24 & +17M \\
SlowFast-NL,16$\times$8~\citep{feichtenhofer2019slowfast} & 234 & 59.9M \\
+\textbf{\ours} & +0.26 & +21.8M \\
\bottomrule
\end{tabular}
\caption{Computational complexity analysis for inference, considering 4 actors and 25 objects per clip.}
\label{tab:FLOPs_Params}
\end{table}


\begin{figure}[t]
\centering
\scriptsize
\setlength{\tabcolsep}{.15em}
\scriptsize
\begin{tabular}{cccc}
Answer phone & Ride (e.g., a bike, a horse) & Take a photo\\
\includegraphics[height=0.250\linewidth]{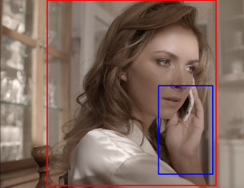} &
\includegraphics[height=0.250\linewidth]{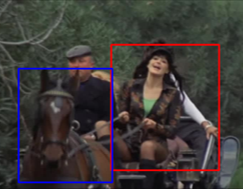} &
\includegraphics[height=0.250\linewidth]{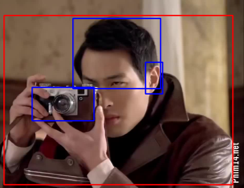} \\
Play musical instrument & Carry/hold (an object) & Kiss (a person) \\
\includegraphics[height=0.250\linewidth]{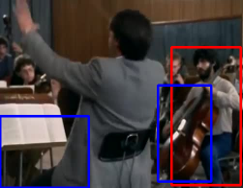} &
\includegraphics[height=0.250\linewidth]{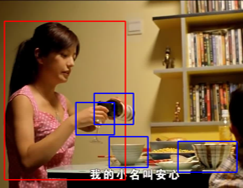} &
\includegraphics[height=0.250\linewidth]{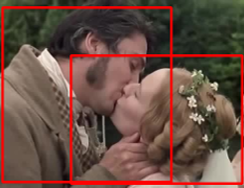} \\
\end{tabular}
\caption{Qualitative results showing the keyframes of the evaluated clips, with red boxes denoting actors performing actions, and blue boxes denoting objects.}
\label{fig:qualitatives}
\vspace{-.3cm}
\end{figure}

\begin{figure}[t]
\centering
\scriptsize
\setlength{\tabcolsep}{.15em}
\begin{tabular}{ccc}
Smoke & Drive (e.g., a car, a truck) & Sail boat\\
\includegraphics[height=0.250\linewidth]{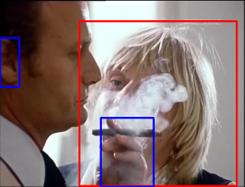} &
\includegraphics[height=0.250\linewidth]{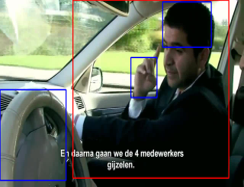} &
\includegraphics[height=0.250\linewidth]{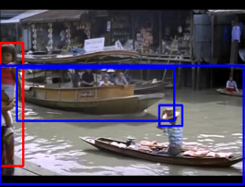} \\
\end{tabular}
\caption{Sample failure cases. Actions are sometimes assigned to an actor very close to the one actually performing the action, and some object-interaction classes are wrongly assigned to people very close to the objects.}
\label{fig:failure}
\vspace{-.3cm}
\end{figure}

\subsection{Qualitative analysis}
\label{sub:qualitatives}
We present some qualitative results obtained on clips of the AVA validation set in Fig.~\ref{fig:qualitatives}. Here, we only show the central keyframe of the clip; red and blue boxes represent predicted actors and objects respectively. For simplicity, we highlight only the actor involved in the action (despite other actors could be found in the scene), except for the \textit{Kiss} class, where two actors perform the same action. Only predicted objects with score greater than $0.8$ are shown, even if all detections are used during training. Fig.~\ref{fig:failure} shows sample failure cases.
On average, we qualitatively observe that our spatio-temporal graph-based module is able to improve the recognition of human actions, especially for actions involving relationships with objects and other people.

\section{Conclusion}
\label{sec:conclusion}
In this work, we presented a novel graph-attention module that can be easily integrated into any video understanding backbone. The module computes updated actor features based on neighboring entities in both space and time. This is done considering detections from consecutive clips as a single learnable graph, where actors and objects are the nodes, while edges hold their relationships. The temporal distance between entities determines the presence of a direct edge between them, while the spatial distance and attention weights define its strength. \textcolor{black}{Through extensive experiments on the Atomic Visual Actions (AVA) dataset and comparisons on J-HMDB-21 and UCF101-24, we showed that our module can bring state-of-the-art performances on a variety of backbones, thus highlighting the role of modeling high-level interactions in both space and time.}

\small{\section*{Acknowledgments}
\vspace{-.2cm}
This work has been partially funded by Metaliquid Srl, Milan (Italy). We also acknowledge NVIDIA AI Technology Center for providing technical support and computational resources.}

\bibliographystyle{model2-names}
\bibliography{egbib}

\end{document}